\def\year{2022}\relax
\documentclass[letterpaper]{article} 
\usepackage{aaai22}  
\usepackage{times}  
\usepackage{helvet}  
\usepackage{courier}  
\usepackage[hyphens]{url}  
\usepackage{graphicx} 
\usepackage{comment} 
\usepackage{natbib}  
\usepackage{caption} 
\DeclareCaptionStyle{ruled}{labelfont=normalfont,labelsep=colon,strut=off} 
\usepackage{algorithm}
\usepackage{algorithmic}

\usepackage{amsmath}
\usepackage{amsthm}
\newtheorem{definition}{Definition}
\usepackage{amsfonts}
\usepackage{stix}
\usepackage{booktabs}

\usepackage{newfloat}
\usepackage{listings}
\floatstyle{ruled}
\newfloat{listing}{tb}{lst}{}
\floatname{listing}{Listing}

\nocopyright

\title{PPFS: Predictive Permutation Feature Selection}

\title{PPFS: Predictive Permutation Feature Selection}
\author {
    Atif Hassan,\textsuperscript{\rm 1}
    Jiaul H. Paik, \textsuperscript{\rm 1}
    Swanand Khare, \textsuperscript{\rm 1}
    Syed Asif Hassan \textsuperscript{\rm 2}
}
\affiliations {
    \textsuperscript{\rm 1} Indian Institute of Technology Kharagpur\\
    \textsuperscript{\rm 2} King Abdulaziz University\\
    atif.hit.hassan@kgpian.iitkgp.ac.in, jia.paik@gmail.com, srkhare@maths.iitkgp.ac.in, shassan1@kau.edu.sa
}

\usepackage{bibentry}

\begin{document}

\maketitle

\begin{abstract}

We propose Predictive Permutation Feature Selection (PPFS), a novel wrapper-based feature selection method based on the concept of Markov Blanket (MB). Unlike previous MB methods, PPFS is a universal feature selection technique as it can work for both classification as well as regression tasks on datasets containing categorical and/or continuous features. We propose Predictive Permutation Independence (PPI), a new Conditional Independence (CI) test, which enables PPFS to be categorised as a wrapper feature selection method. This is in contrast to current filter based MB feature selection techniques that are unable to harness the advancements in supervised algorithms such as Gradient Boosting Machines (GBM). The PPI test is based on the knockoff framework and utilizes supervised algorithms to measure the association between an individual or a set of features and the target variable. We also propose a novel MB aggregation step that addresses the issue of sample inefficiency. Empirical evaluations and comparisons on a large number of datasets demonstrate that PPFS outperforms state-of-the-art Markov blanket discovery algorithms as well as, well-known wrapper methods. We also provide a sketch of the proof of correctness of our method. Implementation of this work is available at \url{https://github.com/atif-hassan/PyImpetus}
\end{abstract}

\section{Introduction}

The primary goal of feature selection is to select relevant subset of features such that they retain the relevant information embedded in the original feature set. Algorithms for feature selection are divided into three categories namely, filter methods, wrapper methods and embedding methods. Filter methods are model-agnostic techniques for feature selection. They do not depend on any predictor for producing output. Filter methods usually work by first establishing an importance measure of a feature with the target variable. Many measures have been defined in literature such as correlation \cite{guyon2002gene,hall2000correlation}, distance \cite{lazzerini2002feature,wolf2005feature}, information such as Mutual Information (MI) \cite{peng2005feature,brown2009new}, consistency \cite{dash2003consistency,robnik2003theoretical} and dependency \cite{jensen2004semantics}. The best set of features are selected based on a comparison of each feature with respect to a given measure.

Unlike filter based feature selection techniques, wrapper methods employ supervised algorithms to greedily evaluate different subsets of the feature space. The subset of features that maximize/minimize a performance metric of said supervised algorithm is produced as output. Wrapper methods are computationally expensive when compared against filter methods but are known to outperform the latter \cite{das2001filters,lee2020markov}. There are many types of wrapper based feature selection techniques such as beam search \cite{siedlecki1993automatic}, randomized hill climbing \cite{skalak1994prototype} and genetic algorithms \cite{holland1992adaptation}. By far the more famous methods are stepwise selection \cite{park2008stepwise} and permutation importance based feature selection \cite{altmann2010permutation}. Stepwise selection, also known as bi-directional elimination, uses a two-step approach to find the best subset of features. It first performs a forward pass in which individual features are passed to a supervised model that provides a score. The feature with the best score is selected. It then performs a backward pass in which the selected subset of features are passed to the same model and are individually eliminated if the error is too large. Both forward and backward passes are computed over multiple iterations. On the other hand, permutation importance based feature selection has a single phase. It first applies a supervised algorithm (model) on the entire dataset and computes a reference score. Then, a single feature is shuffled multiple times and each time a model is used to get a score. The difference between the reference score and the average corrupted scores becomes the importance of that feature. This process is applied on each feature, providing feature importance scores. Finally, features with the best scores are selected.


Due to the high complexity of the feature subset selection problem, most of the feature selection techniques such as stepwise selection and permutation importance follow a greedy method and select the best feature at every round. But there is no guarantee that the set of individually best performing features actually provide the best result. At the same time, methods such as stepwise selection are computationally expensive and consider only a subset of features at each iteration instead of the whole set. Under certain assumptions, borrowing the idea of Markov Blanket (MB) from Bayesian Networks (BN) alleviates these problems. The MB of a node in a BN is the set of all parents, children and spouses (parents of common child) of that node. In terms of feature selection, the MB of a target variable is the set of all features such that the remaining set of features are conditionally independent of the target given its MB. This Conditional Independence (CI) is found by using statistical tests. The MB of a target variable is the minimal subset of features that can provide optimal accuracy~\cite{tsamardinos2003algorithms}. Some of the more famous, theoretically sound MB based feature selection algorithms are the Grow-Shrink (GS) algorithm \cite{margaritis1999bayesian} and the Incremental Association Markov blanket (IAMB) \cite{tsamardinos2003algorithms}.

\subsection{Our Contribution}
All of the MB discovery algorithms have been traditionally recognized as filter methods \cite{fu2010markov}. They are unable to utilize recent improvements in supervised algorithms such as advanced gradient boosting algorithms like XGBoost \cite{chen2016xgboost} and LightGBM \cite{ke2017lightgbm}. At the same time, most of the MB discovery algorithms require different CI tests for different datatypes and tasks. We thus propose a Markov blanket based novel wrapper feature selection algorithm, termed, Predictive Permutation Feature Selection (PPFS). We also propose a novel, wrapper-based non-parametric CI test which we term as Predictive Permutation Independence (PPI) test that can take advantage of recent improvements in supervised algorithms and produce highly accurate Markov blankets for a given dataset. The PPI test allows PPFS to work on datasets containing all types of features including categorical, continuous and mixed. It also allows our method to run on both classification and regression tasks making our method a universal feature selector. A sketch of the proof of the correctness of our algorithm is also provided. To the best of our knowledge, this is the first attempt at developing a Markov blanket based wrapper feature selection algorithm.

\section{Proposed Work}
\subsection{Preliminaries}
\subsubsection{Bayesian Network and Markov Blanket}
A Bayesian network (BN) is a probabilistic graphical model representing the joint probability distribution over a set of features in the form of a Directed acyclic Graph (DAG). The nodes of the graph form the features and the directed edges form the conditional dependencies between these features. If a BN is faithful then the set of parents, children, and spouses (parents of common children) of a node is the unique Markov blanket of that node. 
It has been theoretically proven that the MB of a target node is the minimal set of nodes that can sufficiently explain the target \cite{koller1996toward}. Therefore, in terms of feature selection, under the assumption of faithfulness and the existence of a correct conditional independence test, the Markov blanket of a target is the unique, minimal set of features that can sufficiently explain the target variable.

\subsubsection{Conditional Independence Test}
A Conditional Independence (CI) test is required in order to check whether two variables are independent of each other conditioned on a separate set of features. Let $X$ and $Y$ be two variables whose independence needs to be checked and let $Z$ be the conditioning set. The CI test corresponds to checking if $P(X,Y | Z) = P(X | Z) P(Y | Z)$. For categorical features, $G^2$ or \raisebox{2pt}{$\chi^2$} tests are used~\cite{pena2007towards}, while for continuous features Student's t-test or the Fisher's test is used~\cite{pena2007towards}. It is important to point out that until the proposal of Mixed-MB~\cite{lee2020markov}, Markov blanket based feature selection algorithms required different CI tests for different feature types.

\subsubsection{Basic Notations and Definitions}
Let $X$ denote a dataset in $\mathbb{R}^{n \times d}$ where $n$ is the number of samples in the dataset and $d$ the dimensionality. We denote the set of all features in $X$ by $S$ such that $S=\{X_1, X_2, ... X_d\}$ where $X_i$ is an individual feature. A single sample from $X$ is written as $x_i$. $\widetilde{X}$ denotes permuted version of $X$ where only one feature is permuted. Let $Y$ be the target variable. A single sample from the target variable is written as $y_i$. The Markov blanket of $Y$ is denoted as $\text{MB}(Y)$. Conditional Independence (CI) is denoted by $\Vbar$, that is, if $X$ is conditionally independent of $Y$ given some variable $Z$, then it is written as $(X \Vbar Y \; | \; Z)$. The null hypothesis of the CI test is denoted by $H_0$.

\begin{definition}
    \textbf{(Markov Condition)} Any variable (node) in a bayesian network is independent of its non-descendants given its parents \cite{margaritis2003learning}.
\end{definition}

\begin{definition}
    \textbf{(Faithfulness)} Let $G$ denote a Bayesian network. Let $P$ denote a joint probability. $B$ and $P$ are said to be faithful to one another iff all the conditional independencies entailed by $G$ and the Markov condition is present in $P$ \cite{fu2010markov}.
\end{definition}

\begin{definition}
    \textbf{(Markov Blanket)} Under the faithful condition, $\text{MB}(Y)$ is the minimal set conditioned on which all other variables are independent of $Y$, i.e., $(X \setminus MB(Y) \Vbar Y \; | \; MB(Y))$ \cite{margaritis1999bayesian}.
\end{definition}

\subsection{Predictive Permutation Independence Test}
\label{sec:ppi}
The Predictive Permutation Independence (PPI) test is a novel, non-parametric generalized Conditional Independence (CI) test that can take advantage of recent advancements in supervised algorithms such as gradient boosting machines to correctly uncover underlying joint probability distributions. The PPI test is built on the popular knockoff framework \cite{candes2016panning} and is similar to the Conditional Predictive Impact test \cite{watson2019testing}. The main idea behind the knockoff framework is that if a feature is unable to perform significantly better than its knockoff counterpart by some importance measure then the feature is not important.

Let $X$ denote a feature matrix in $\mathbb{R}^{n \times d}$ where $n$ is the number of samples in the dataset and $d$ the dimensionality. Let $S=\{X_1, X_2 \; ... \; X_d\}$ be the set of all features in the dataset such that $X_i \in X$ be a single feature. A single sample/row in $X$ is denoted by $x_i$. Let $Y$ be the target vector of size $n \times 1$ and $y_i$ be a single sample in $Y$. The feature matrix and the target variable taken together form dataset $Z=(X,Y)$. The null hypothesis of PPI is defined as,
\begin{equation}
\label{null_hypo_ppi}
    H_0: X_i \Vbar Y \; | \; U
\end{equation}
where $U \subseteq S \setminus X_i$. Therefore, $U$ can also be an empty set which is the case for marginal independence.

We first generate $B$ copies/duplicates of $Z$. Here $B$ is a hyper-parameter. Let $Z^j$ denote a single copy where $j=\{1,2,..\;B\}$. A random train-test (80\%-20\%) split is performed on each $Z^j$ giving rise to $Z^j_{\text{train}}$ and $Z^j_{\text{test}}$. Typical values of $B$ lie within the range $5 \leqslant B \leqslant n$, where $n$ is the number of samples while $5$ comes from the fact that we consider 80\%-20\% train-test split resulting in a minimum $\frac{n}{0.2n}$ required splits. Let $f: X \rightarrow Y$ be a predictive function that learns a mapping from feature space to output. This $f$ can be any supervised learning algorithm. Since there are $B$ copies of the data, therefore, $B$ separate functions need to learned. Each function is denoted by $f^j$ where $j=\{1,2,..\;B\}$. Let $L: \left( \widehat{Y}, Y \right) \rightarrow \mathbb{R}^n$ be a loss function that maps predicted output, $\widehat{Y}$, and actual output, $Y$, to an error vector of size $n$. Since there are $B$ copies of the data, therefore, $B$ separate error vectors will be generated, denoted by $L\left(\widehat{Y^j}, Y^j  \right)$. Here, $\widehat{Y^j}$ and $Y^j$ correspond to the predicted and actual output of $Z^j$, respectively. For classification and regression tasks, $L$ is chosen as log-loss and mean squared error respectively. Once each $f^j$ has been trained on corresponding $Z^j_{\text{train}}$, the empirical risk over each $Z^j_{\text{test}}$ can be written as,
\begin{equation}
    \label{loss}
    R^j = \frac{1}{n} \left\lVert L\left( f^j(X_{\text{test}}^j), Y_{\text{test}}^j \right)\right\rVert_1
\end{equation}

In order to check the importance of a feature, it needs to be compared with its knockoff. The knockoff is generated by permuting the feature. This breaks the association of that feature with the target variable as the joint probability of the feature and the target changes. Let $X_i$ be the feature that needs to be checked for conditional independence. Then only $X_{i_{\text{test}}}^j$ in each corresponding $Z_{\text{test}}^j$ is permuted. Here permutation means shuffling the rows of said feature. Let the knockoff test sets be denoted as $\widetilde{Z_{\text{test}}^j}=(\widetilde{X_{\text{test}}^j},Y_{\text{test}}^j)$. Then the empirical risk over each $\widetilde{Z_{\text{test}}^j}$ can be written as,
\begin{equation}
    \label{emp_risk}
    \widetilde{R^j} = \frac{1}{n} \left\lVert L\left( f^j(\widetilde{X_{\text{test}}^j}), Y_{\text{test}}^j \right)\right\rVert_1
\end{equation}
Let $R$ be a variable comprising of $R^j$ and $\widetilde{R}$ be a variable comprising of $\widetilde{R^j}$. We can then replace the null hypothesis, as defined in Equation~\ref{null_hypo_ppi} by,
\begin{equation}
\label{new_null_hypo_ppi}
    H_0: \mathbb{E}(R) \geq \mathbb{E}(\widetilde{R})
\end{equation}
This means that the expected risk of a model with a given feature is greater than the expected risk of that model with the feature's knockoff counterpart. The null hypothesis in Equation~\ref{new_null_hypo_ppi} can be verified by any non-parametric, one-tailed, paired significance test. We use the Wilcoxon's signed rank test for this purpose. If the p-value is less than some significance level, say $\alpha$ then the null hypothesis can be rejected meaning that the feature is important as it decreases the risk associated with a model. Algorithm \ref{algo:PPI} provides pseudocode for the proposed test.

\begin{algorithm}[ht]
\caption{Psuedocode for PPI test}
\label{algo:PPI}
\textbf{Input}: Feature Matrix $X$; Target Variable $Y$; Conditioning Set $Q$; Feature index $i$; Number of copies $B$; Supervised model $f$; Loss function $L$; Non-parametric one-tailed paired significance test $h$;\\
\textbf{Output}: Gives the significance score of the feature at input index $i$
\begin{algorithmic}[1] 
\STATE $X \gets \text{Concatenate}(X, Q)$
\STATE $Z = []$
\FOR{$j \gets 0$ to $B$}
\STATE $Z[j] \gets (X,Y)$
\ENDFOR
\STATE $R = []$
\STATE $\widetilde{R} = []$.

\FOR{$j \gets 0$ to $B$}
\STATE  $Z_{\text{train}}, Z_{\text{test}} = \text{random\_train\_test\_split(Z)}$
\STATE train($f(Z_{train})$)
\STATE $X_{\text{test}}, Y_{\text{test}} \gets Z_{\text{test}}$
\STATE $R[j] \gets \frac{1}{n}\left\lVert L \left(f \left(X_{\text{test}} \right), Y_{\text{test}} \right)\right\rVert_1$
\STATE $\widetilde{X_{\text{test}}} \gets \text{shuffle}(X_{\text{test}}[i])$
\STATE $\widetilde{R[j]} \gets \frac{1}{n}\left\lVert L \left(f \left(\widetilde{X_{\text{test}}} \right), Y_{\text{test}} \right)\right\rVert_1$
\ENDFOR
\STATE \textbf{return} $h(R, \widetilde{R})$
\end{algorithmic}
\end{algorithm}


\subsection{Predictive Permutation Feature Selection}
\label{sec:ppfs}
Similar to GS and IAMB, Predictive Permutation Feature Selection (PPFS) employs a growth phase and a shrink phase. The growth phase includes features into a candidate MB using marginal independence by conditioning on the empty set. The shrink phase eliminates false positives from the candidate MB using conditional independence to finally recommend an optimal subset of features. We identify that removal of a feature in the shrink phase leads to change in the joint probability distribution over the candidate MB. Ideally, after removal of a single feature, the shrink phase should restart in order to take into account the new joint probability distribution but such a method would be computationally expensive. In order to tackle this issue, PPFS performs a simple heuristic based ordering over the features in the candidate MB at the end of growth phase. PPFS also employs a novel K-Fold aggregation step for small datasets as well as datasets that do not follow the faithfulness assumption in order to address the issue of sample inefficiency. The entire MB aggregated version of PPFS algorithm is provided in Algorithm \ref{algo:PPFS}.

\subsubsection{Growth Phase}
The growth phase of PPFS determines which features are individually important to predict the target variable. Let $X \in \mathbb{R}^{n \times d}$ be an input feature matrix. Let $S$ be the set of features in $X$ such that, $S=\{X_1, X_2, ... \; X_d\}$. Let $Y$ be the target variable and $\text{MB}(Y)$ be the Markov Blanket of $Y$. In order to determine individual importance, the PPI test checks for marginal independence by conditioning each feature, $X_i \in S$, on the empty set. The test returns a p-value which if less than some significance level $\alpha$ results in the addition of that feature into MB$(Y)$ since the feature is important for predicting the target variable. This process is shown in Eqn. \ref{growth_phase}.
\begin{equation}
\begin{split}
    \text{MB}(Y) = \{X_i \in S \; \; | \; \; \text{PPI}(X_i \Vbar Y \; | \; \emptyset) < \alpha, \\
    \forall \; i \in \{1,2 \; ... \; d\} \}
\end{split}
    \label{growth_phase}
\end{equation}


\subsubsection{Shrink Phase}
The shrink phase is responsible for removing any false positives that might have been added to the MB during the growth phase. Each feature, $X_i \in \text{MB}(Y)$ is checked for conditional independence, using the PPI test, conditioned on the rest of the Markov Blanket, $\text{MB}(Y) \setminus \{X_i\}$. If the value from the PPI test is not significant then the feature, $X_i$ is removed from $\text{MB}(Y)$ and the Markov Blanket is updated as the feature is not important with respect to the remaining $\text{MB}(Y)$. Once this happens, the entire shrink phase begins again and the process continues until no more features can be removed during a single iteration of the shrink phase.


\subsubsection{Improved Shrink Phase}
Removing a feature from the candidate MB set, changes the underlying joint probability distribution of the Markov blanket. In order to capture the change, shrink phase is restarted. This process quickly becomes computationally expensive. In order to address said issue, we use p-values, generated during the growth phase, as feature importance scores to rank the features selected in the Markov blanket before starting the shrink phase. Sorting features in ascending order of their importance allows the shrink phase to remove false positives before arriving at the more important features. Thus, the more important features are not removed before the less important features.
\begin{definition}
    \textbf{(Feature Importance Score)} Under the assumption that the p-value returned by the PPI test is correct and that smaller p-values denote higher importance, the feature importance score can be defined as $log(\frac{1}{\text{p-value}})$. Assuming that the largest admissible significant p-value is $0.05$, the range of this proposed score is $[2.99,\; \infty)$.
\end{definition}

Thus, once the candidate MB set from the growth phase is sorted in increasing order of importance, the shrink phase is initiated. If a feature is found to be a false positive, it is removed from the Markov Blanket with $\text{MB}(Y)$ being updated accordingly and the iteration continues. This is in contrast to the earlier version of the shrink process wherein the entire phase was restarted upon deletion of each false positive from $\text{MB}(Y)$.

\begin{algorithm}[ht]
\caption{Psuedocode for PPFS with MB aggregation}
\label{algo:PPFS}
\textbf{Input}: Feature Matrix $X$; Target Variable $Y$; Number of subsets $K$ for cross-validation; Significance threshold $\alpha$\\
\textbf{Output}: Selects the best MB
\begin{algorithmic}[1] 
\STATE all\_MB = []
\STATE new\_$X$, new\_$Y \gets \text{split}(X, Y, K)$\;
\FOR{$k \gets 0$ to $K$}
\STATE $X, Y \gets \text{new\_}X[k], \; \text{new\_}Y[k]$.
\STATE MB, $\;$ p\_values, $\;$ $j \gets$ [ ], $\;$ [ ], $\;$ $0$

\FOR{$i \gets 0$ to $d$}
\STATE $p \gets \text{PPI}(X_i \Vbar Y \; | \; \emptyset)$.
\IF {$p \leq \alpha$}
\STATE $\text{MB} \gets \text{MB} \cup {X_i}$
\STATE p\_values[$j$] $\gets$ p
\STATE $j$++
\ENDIF
\ENDFOR

\STATE MB$ \gets \text{sort}(\text{MB}, \; \text{p\_values})$
\FOR{$i \gets 0$ to $m$}
\IF{\text{PPI}$(X_i \Vbar Y \;|\; \text{MB} \setminus \{X_i\}) > \alpha$}
\STATE $\text{MB} = \text{MB} \setminus \{X_i\}$
\ENDIF
\ENDFOR

\STATE all\_MB[$k$] $\gets$ MB
\ENDFOR

\STATE Follow equations \ref{feature_aggregation}, \ref{feature_score} and \ref{MB_score} to get array $Z$
\STATE $i$ $\gets$ argmax($Z$)
\STATE \textbf{return} all\_MB[$i$]
\end{algorithmic}
\end{algorithm}

\subsubsection{Markov Blanket Aggregation}
Similar to algorithms such as GS, IAMB and some of its variants, PPFS is also sample inefficient. Reason being that the growth phase is conditioned on the empty set. This leads to the addition of a lot of false positives resulting in a large candidate MB. Since sample requirement is exponential with respect to the number of features, therefore, PPFS is sample inefficient. This is not an issue for datasets with large sample sizes, though, our algorithm might provide incorrect results for very small datasets. At the same time, if a dataset violates the faithful assumption, PPFS again might not provide the correct Markov blanket since there does not exist a unique MB$(Y)$. In order to address these shortcomings, we propose a novel Markov blanket aggregation step that is treated as a hyper-parameter and can be switched on or off by the user.

We use a K-Fold MB aggregation strategy in our PPFS algorithm. The data is first divided into K separate folds. Both growth and shrink phases are run on each fold, sequentially, which yields K separate Markov Blankets. Then, a new set, $Q$, is generated which is the union of all the K different MBs as shown in Eqn. \ref{feature_aggregation}
\begin{equation}
    \label{feature_aggregation}
    Q = \bigcup\limits_{i=1}^{K} \text{MB}_i
\end{equation}
For each feature in $Q$, its corresponding frequency of occurrence in every Markov blanket is calculated as shown in Equation~\ref{feature_score}
\begin{equation}
    \label{feature_score}
    \begin{split}
        \text{freq}(X_j) = \sum_{i=1}^{K}F_i, \;\; \text{where}\\
        F_i =
            \begin{cases}
            1 \;\; \text{if} \;\; X_j \in \text{MB}_i(Y)\\
            0 \;\; \text{otherwise}
            \end{cases}
    \end{split}
\end{equation}
Here, $X_j \in Q$ and $j \in \{1, 2 \; ... \; |Q|\}$. Let $Z$ be a list containing scores, $z_i$, calculated for each Markov Blanket. Here, $i \in \{1, 2 \; ... \; K\}$. Then,
\begin{equation}
    \label{MB_score}
    z_i = \frac{\sum_{j=1}^{|\text{MB}_i(Y)|}\text{freq}(X_j)}{|\text{MB}_i(Y)|}
\end{equation}
Since Markov blankets are the minimal set of features that can predict the target variable, the scores calculated for each MB is normalized by the number of features it contains (Equation~\ref{MB_score}). The final MB is chosen as the one with the highest $z_i$ value. 
Intuitively, $z_i$ scores reflect the 
potential of a Markov blanket as a good quality predictor and the score is higher if the features of that Markov blanket appear in many other Markov blankets generated from the $K$ folds.

\subsection{Sketch of Proof of Correctness}
The Markov blanket of a target variable consists of its parent, child and spouse features. The parent and child are strongly associated with the target variable and do not require any conditioning set. Hence, such features will be inducted into the candidate MB during the growth phase. Spouses are weakly associated with the target variable in the absence of the common child in the conditioning set \cite{tsamardinos2003algorithms}. In order to prove that PPFS's growth phase also inducts spouses into the candidate set, we utilize the permutation property of the PPI test. Let $g: (X, \widetilde{X}) \rightarrow \mathbb{R}$ be a function that takes two datasets as input and outputs a score which measures the domain shift between them. Clearly, $g(X_{i_{\text{train}}}, X_{i_{\text{test}}}) < g(X_{i_{\text{train}}}, \widetilde{X_{i_{\text{test}}}})$ where, $X_{i_{\text{train}}}$ and $X_{i_{\text{test}}}$ are the train and test splits of feature $X_i \in X$ while $\widetilde{X_{i_{\text{test}}}}$ is $X_{i_{\text{test}}}$ but permuted. Here $X$ is a feature matrix and $X_i$ is a spouse. Keeping in mind that spouses have at least some association with the target variable and the fact that the error of a supervised method increases with increase in domain shift \cite{tzeng2014deep}, we can say that $\mathbb{E}(R) < \mathbb{E}(\widetilde{R})$ where $R$ is the empirical risk over $X_{i_{\text{test}}}$ while $\widetilde{R}$ is the empirical risk over $\widetilde{X_{i_{\text{test}}}}$. Thus, the null hypothesis in Equation~\ref{new_null_hypo_ppi} can be rejected which proves that spouses are also added into the candidate MB set during growth phase. This consequently proves that the candidate MB set after the growth phase is a superset of the actual MB$(Y)$. Under the faithfulness assumption and the Markov condition, the shrink phase will remove all false positives that are not either a parent, a child or a spouse of the target variable in the underlying BN, conditioned on MB$(Y)$. Thus, after the shrink phase, PPFS returns a unique and correct MB$(Y)$.

\section{Experimental Setup}
\subsection{Datasets}
We compare PPFS against two state-of-the-art MB discovery algorithms, Mixed-MB \cite{lee2020markov} and SGAI \cite{yu2016markov}. We also provide comparison against the two famous wrapper methods, stepwise selection and permutation importance based feature selection. We consider a total of $12$ datasets for verifying the efficacy of our proposed methodology. These include $8$ datasets from the UCI machine learning repository comprising of Statlog \cite{brown2004diversity}, Iono \cite{sigillito1989classification}, Sonar \cite{gorman1988analysis}, WBDC \cite{street1993nuclear}, Credit \cite{quinlan1987simplifying}, Ecoli \cite{horton1996probabilistic}, Desharnais \cite{desharnais1989analyse} and Maxwell \cite{maxwell2002applied}. The remaining $4$ are from the NIPS $2003$ feature selection challenge comprising of Arcene \cite{guyon2004result}, Dexter \cite{guyon2004result}, Dorothea \cite{guyon2004result} and Madelon  \cite{guyon2004result} datasets. Together, these datasets provide a real-world, exhaustive empirical evaluation of our method in comparison to other feature selection techniques. Table \ref{table:data} provides a detailed view of all the datasets. Here BC stands for Binary Classification, MC for Multi Classification and R for Regression

\begin{table}[ht]
  \caption{Description of the datasets used for evaluation. Here, BC stands for Binary Classification, MC for Multi Classification and R for Regression}
  \label{table:data}
  \begin{tabular}{lllll}
    \toprule
    Dataset & Features & Samples & Classes & Task\\
    \midrule
    Statlog  & 13 & 270 & 2 & BC\\
    Iono & 33 & 351 & 2 & BC\\
    Sonar & 60 & 208 & 2 & BC\\
    WDBC & 30 & 569 & 2 & BC\\
    Credit & 15 & 653 & 2 & BC\\
    Ecoli & 7 & 336 & 7 & MC\\
    Desharnais & 11 & 81 & NA & R\\
    Maxwell & 26 & 62 & NA & R\\
    Arcene & 10000 & 100 & 2 & BC\\
    Dexter & 20000 & 300 & 2 & BC\\
    Dorothea & 100000 & 800 & 2 & BC\\
    Madelon & 500 & 2000 & 2 & BC\\
  \bottomrule
\end{tabular}
\end{table}

\subsection{Baselines}
Mixed-MB utilizes a likelihood ratio based generalized CI test for discovering the MB of a target variable on any given mixed-type dataset. SGAI (Selection via Group Alpha-Investing) on the other hand is a state-of-the-art MB discovery algorithm under the non-faithful assumption. It takes a different approach to finding MB of a target variable using the concept of representative sets. This allows SGAI to find the optimal MB without exhaustive search over all other possible MBs under the non-faithful condition. Additionally, we compare our methods against the well known wrapper methods, Stepwise Feature Selection (SFS) \cite{park2008stepwise} and Permutation Importance based Feature Selection (PIFS) \cite{altmann2010permutation}. SFS uses forward and backward phases at each iteration for selecting candidate features. PIFS uses multiple shuffled versions of each feature to compute importance scores that are then utilized to rank and select the candidate set.

\section{Results}
In order to compare our method against Mixed-MB and SGAI, the experimental settings were chosen to be as close to each of the works as possible. For Mixed-MB, $5$-fold cross validation is used with both decision tree and SVM classifiers. The performance is reported on classification accuracy. For SGAI, $10$-fold cross validation is used with SVM classifier and the performance is reported on prediction error. The model supplied to the PPI test in all experiments is a decision tree. The value of $B$ and $K$ varies for each dataset and are decided based on nested cross-validation \cite{yu2016markov}. Their values for each experiment are present in corresponding tables for reproducibility. The significance level, $\alpha$ is set to $0.05$ for all experiments. For SVM classifier, we use rbf kernel and set $C=1$.  All experiments are conducted on python using scikit-learn library \cite{scikit-learn}. All experiments were run on a windows laptop with $16$gb ram and a $10$-th gen core i7.

\subsection{Performance on Classification Tasks}
Table~\ref{table:Table_1} reports the performance of the methods on datasets with small number of samples and features. Decision Tree is used for classification for all methods on the selected features as well as for the selection of the candidate set for PIFS and SFS. The table shows that the proposed model has outperformed the baseline models. The features selected by all the models improve the classification performance compared to classification that includes all the features.

\begin{table}[ht]
    \setlength{\tabcolsep}{4.5pt}
  \caption{Comparison of PPFS against PIFS, SFS and Mixed-MB (MMB) with decision tree classifier as the model and classification accuracy as metric on datasets with small number of features. Here All denotes model trained on all features. The values for hyperparameters B and K are also provided for each dataset.}
  \label{table:Table_1}
  \begin{tabular}{llllllll}
    \toprule
    Dataset & All & PIFS & SFS & MMB & PPFS & B & K\\
    \midrule
    Statlog & 0.711 & 0.739 & 0.742 & 0.773 & \textbf{0.837} & 30 & 5\\
    Iono & 0.880 & 0.884 & 0.897 & 0.884 & \textbf{0.923} & 10 & 0\\
    Sonar & 0.716 & 0.751 & 0.754 & \textbf{0.807} & 0.784 & 10 & 5\\
    WDBC & 0.933 & 0.922 & 0.937 & 0.928 & \textbf{0.949} & 50 & 0\\
    Credit & 0.831 & 0.842 & 0.849 & 0.846 & \textbf{0.864} & 50 & 0\\
    Ecoli & 0.807 & 0.805 & 0.816 & 0.813 & \textbf{0.822} & 10 & 0\\
  \bottomrule
\end{tabular}
\end{table}

Table~\ref{table:Table_2} compares the performance of the methods with SVM as the classifier. SVM is also used for the selection of the candidate set for PIFS and SFS. Similar to Table~\ref{table:Table_1}, PPFS outperforms all other methods. In this case too, the feature selection improved the performance of the downstream classification model. On the selected features, SVM does better than the decision tree.

\begin{table}[ht]
    \setlength{\tabcolsep}{4.5pt}
  \caption{Comparison of PPFS against PIFS, SFS and Mixed-MB (MMB) with SVM classifier as the model and classification accuracy as metric on datasets with small number of features. Here All denotes model trained on all features. The values for hyperparameters B and K are also provided for each dataset.}
  \label{table:Table_2}
  \begin{tabular}{llllllll}
    \toprule
    Dataset & All & PIFS & SFS & MMB & PPFS & B & K\\
    \midrule
    Statlog & 0.833 & 0.839 & 0.840 & \textbf{0.848} & 0.844 & 30 & 5\\
    Iono & 0.940 & 0.940 & 0.943 & 0.906 & \textbf{0.946} & 50 & 0\\
    Sonar & 0.851 & 0.851 & 0.852 & 0.783 & \textbf{0.856} & 30 & 5\\
    WDBC & 0.971 & 0.971 & 0.973 & 0.953 & \textbf{0.979} & 50 & 0\\
    Credit & 0.860 & 0.862 & 0.865 & 0.859 & \textbf{0.867} & 50 & 0\\
    Ecoli & 0.863 & 0.869 & 0.871 & 0.872 & \textbf{0.875} & 30 & 0\\
  \bottomrule
\end{tabular}
\end{table}

Table~\ref{table:Table_3} reports the results of the models on datasets with large number of features. Only SVM classifier is used for comparing PPFS with SGAI and other methods because SGAI uses SVM as its downstream classification model. SVM is used for the selection of the candidate set for PIFS and SFS. For this experiment, we report the prediction error rather than accuracy, simply because the reported results of the existing methods on these datasets are in terms of prediction error. It is evident that the proposed model unequivocally outperforms all the baseline models on all the datasets.

\begin{table}[ht]
    \setlength{\tabcolsep}{4.3pt}
  \caption{Comparison of PPFS against PIFS, SFS and SGAI with SVM classifier as the model and prediction error as metric on datasets with large number of features. Here All denotes model trained on all features. The values for hyperparameters B and K are also provided for each dataset.}
  \label{table:Table_3}
  \begin{tabular}{llllllll}
    \toprule
    Dataset & All & PIFS & SFS & SGAI & PPFS & B & K\\
    \midrule
    Arcene & 0.180 & 0.172 & 0.170 & 0.190 & \textbf{0.153} & 10 & 5\\
    Dexter & 0.170 & 0.149 & 0.111 & 0.107 & \textbf{0.088} & 5 & 0\\
    Dorothea & 0.091 & 0.085 & 0.773 & 0.060 & \textbf{0.043} & 5 & 0\\
    Madelon & 0.351 & 0.217 & 0.191 & 0.189 & \textbf{0.131} & 5 & 0\\
  \bottomrule
\end{tabular}
\end{table}

\subsection{Performance on Regression Tasks}
This section evaluates PPFS against other methods on regression tasks. For these experiments we use two datasets and support vector regression (SVM) as shown by Mixed-MB. Table~\ref{table:Table_4} reports the mean squared error of the models on the two datasets. Similar to classification tasks, PPFS once again is the best performer, while Mixed-MB seems to be the best baseline model.
\begin{table}[ht]
    \setlength{\tabcolsep}{4.0pt}
  \caption{Comparison of PPFS against PIFS, SFS and Mixed-MB (MMB) with SVM as the regression model and mean squared error as metric. Here All denotes model trained on all features. The values for hyperparameters B and K are also provided for each dataset.}
  \label{table:Table_4}
  \begin{tabular}{llllllll}
    \toprule
    Dataset & All & PIFS & SFS & MMB & PPFS & B & K\\
    \midrule
    Desharnais & 3917 & 3877 & 3811 & 3830 & \textbf{2732} & 50 & 5\\
    Maxwell & 7624 & 6702 & 6555 & 6301 & \textbf{5387} & 35 & 5\\
  \bottomrule
\end{tabular}
\end{table}

\subsection{Number of Selected Features}
In previous sections, we focused on the prediction accuracy/errors of the selected features. In this section we compare the number of selected features by each method on their respective datasets. From Tables ~\ref{table:Table_5} and \ref{table:Table_6}, it is clear that PPFS outperforms Mixed-MB and other methods on both large and small datasets by selecting the least number of features. Often the number of selected features are less than $50\%$ of the total data. Even though PPFS achieves high feature reduction compared to the other methods, is still achieves the best prediction performance. The results also clearly show that PPFS is robust as well as scalable with datasets with large number of features.

\begin{table}[ht]
  \caption{Comparison of number of features selected by PPFS against PIFS, SFS and Mixed-MB (MMB) on each dataset. Lesser is better.}
  \label{table:Table_5}
  \begin{tabular}{llllll}
    \toprule
    Dataset & All & PIFS & SFS & MMB & PPFS\\
    \midrule
    Statlog & 13 & 10 & 8 & 8 & 3\\
    Iono & 33 & 27 & 23 & 12 & 14\\
    Sonar & 60 & 32 & 24 & 9 & 7\\
    WDBC & 30 & 22 & 13 & 7 & 9\\
    Credit & 15 & 8 & 7 & 4 & 2\\
    Ecoli & 7 & 6 & 6 & 6 & 6\\
    Desharnais & 11 & 10 & 10 & 2 & 4\\
    Maxwell & 26 & 9 & 6 & 4 & 3\\
    \midrule
    Average & 24.4 & 15.5 & 12.1 & 6.4 & \textbf{6}\\
  \bottomrule
\end{tabular}
\end{table}

\begin{table}[ht]
  \caption{Comparison of number of features selected by PPFS against PIFS and SFS on each dataset. Lesser is better.}
  \label{table:Table_6}
  \begin{tabular}{lllll}
    \toprule
    Dataset & Total & PIFS & SFS & PPFS\\
    \midrule
    Arcene & 10000 & 902 & 459 & 304\\
    Dexter & 20000 & 2100 & 437 & 384\\
    Dorothea & 100000 & 10500 & 5103 & 2001\\
    Madelon & 500 & 210 & 85 & 12\\
    \midrule
    Average & 32625 & 3428 & 1521 & \textbf{675.25}\\
  \bottomrule
\end{tabular}
\end{table}


\section{Related Work}
All Markov blanket based feature selection algorithms assume that the underlying processes generating the data can be represented by a Bayesian network and that there exists reliable statistical CI tests to measure the association of a feature with the target variable. These assumptions allow an MB feature selection algorithm to discover a unique MB for a given data. The Grow-Shrink (GS) algorithm \cite{margaritis1999bayesian} was the first, theoretically proven MB discovery algorithm, under these assumptions. The GS algorithm consists of a growth phase followed by a shrink phase. The features are initially sorted in decreasing order of their association with the target variable. The algorithm then adds a feature to the candidate MB if it is conditionally dependent on the target variable given the MB. After the growth phase, the shrink phase utilizes the same CI test to remove any false positives that might have been added to the MB. Due to the initial ordering of the features, many false positives are added to the candidate MB before the spouses of the target variable. This is because spouse features are weakly associated with the target variable without the common child in the conditioning set. Consequently, the shrink phase becomes unreliable since the number of samples required to get an accurate prediction is exponential with respect to the size of MB.

To solve said issue, the Incremental Association Markov blanket (IAMB) \cite{tsamardinos2003algorithms} was proposed. Unlike GS, IAMB does not sort the features before starting its growth phase. This leads to lesser false positives being added to the candidate MB set which produces the correct MB a lot faster then GS. But IAMB is also sample inefficient. Multiple modified versions of the IAMB algorithm have been proposed over the years such as IAMBnPC \cite{tsamardinos2003algorithms}, inter-IAMBnPC \cite{tsamardinos2003algorithms}, inter-IAMB \cite{tsamardinos2003algorithms} and Fast-IAMB \cite{yaramakala2005speculative}. GS, IAMB and its variants require different CI tests for both classification and regression as well as for different features types such as continuous and categorical. This problem is addressed by the recently proposed state-of-the-art Mixed-MB \cite{lee2020markov} algorithm. It utilizes a likelihood ratio based generalized CI test for discovering the MB of a target variable on a given mixed-type dataset.

GS, IAMB and its variants as well as Mixed-MB are all sample inefficient. To tackle this problem Min-Max MB (MMMB) \cite{tsamardinos2003time} was introduced. The MMMB algorithm divides the problem of MB discovery into two sub-problems. The first sub-problem deals with finding the parent and child of target variable while the second sub-problem deals with finding the spouses. Hiton-MB \cite{aliferis2003hiton} is a modification over the MMMB algorithm. It interleaves the growth and shrink phase in order to further reduce the number of false positives in the candidate MB set.

Even though MMMB and Hinton-MB addressed the issue of sample inefficiency to a certain extent, they were not proved to be theoretically sound. On the other hand, all of the stated algorithms can discover a unique MB only if the dataset can be faithfully represented by a BN. This faithfulness assumption may not always hold and hence all stated algorithms would fail in such a situation. TIE* (Target Information Equivalence) \cite{statnikov2013algorithms} discovers all Markov blankets under the non-faithful condition but becomes computationally infeasible when the number of MBs grow exponentially with respect to the number of features in the dataset. SGAI (Selection via Group Alpha-Investing) \cite{yu2016markov}, a state-of-the-art MB discovery algorithm under the non-faithful assumption takes a different approach to finding MB of a target variable. It proposes the concept of representative sets and uses it to find the optimal MB without exhaustive search over all other MBs.

\section{Conclusion}
To the best of our knowledge, PPFS is the first attempt at proposing a novel Markov blanket based wrapper feature selection technique. We also introduce a new conditional independence test, PPI, based on the knockoff framework. PPFS is a universal feature selection method as it is able to perform subset selection for datasets containing both categorical and continuous features while being applicable to both classification as well as regression tasks. We prove that our method is theoretically sound and under certain assumptions provides the correct unique Markov blanket. Evaluation of PPFS against state-of-the-art Markov blanket discovery algorithms and famous wrapper based feature selection techniques demonstrates the power of our method in terms of metric based performance. Thus, PPFS is a great alternative to other feature selection techniques and can help improve overall performance of any machine learning pipeline. 

\bibliography{aaai22}

\end{document}